\PassOptionsToPackage{unicode}{hyperref}
\PassOptionsToPackage{hyphens}{url}
\documentclass[
]{article}
\usepackage{xcolor}
\usepackage[margin=1in]{geometry}
\usepackage{amsmath,amssymb}
\usepackage{iftex}
\ifPDFTeX
  \usepackage[T1]{fontenc}
  \usepackage[utf8]{inputenc}
  \usepackage{textcomp} 
\else 
  \usepackage{unicode-math} 
  \defaultfontfeatures{Scale=MatchLowercase}
  \defaultfontfeatures[\rmfamily]{Ligatures=TeX,Scale=1}
\fi
\usepackage{lmodern}
\ifPDFTeX\else
\fi
\IfFileExists{upquote.sty}{\usepackage{upquote}}{}
\IfFileExists{microtype.sty}{
  \usepackage[]{microtype}
  \UseMicrotypeSet[protrusion]{basicmath} 
}{}
\makeatletter
\@ifundefined{KOMAClassName}{
  \IfFileExists{parskip.sty}{%
    \usepackage{parskip}
  }{
    \setlength{\parindent}{0pt}
    \setlength{\parskip}{6pt plus 2pt minus 1pt}}
}{
  \KOMAoptions{parskip=half}}
\makeatother
\usepackage{color}
\usepackage{fancyvrb}

\DefineVerbatimEnvironment{Highlighting}{Verbatim}{commandchars=\\\{\}}
\usepackage{framed}
\definecolor{shadecolor}{RGB}{248,248,248}
\newenvironment{Shaded}{\begin{snugshade}}{\end{snugshade}}

\newcommand{\AttributeTok}[1]{\textcolor[rgb]{0.13,0.29,0.53}{#1}}

\newcommand{\BuiltInTok}[1]{#1}
\newcommand{\CharTok}[1]{\textcolor[rgb]{0.31,0.60,0.02}{#1}}
\newcommand{\CommentTok}[1]{\textcolor[rgb]{0.56,0.35,0.01}{\textit{#1}}}

\newcommand{\DataTypeTok}[1]{\textcolor[rgb]{0.13,0.29,0.53}{#1}}
\newcommand{\DecValTok}[1]{\textcolor[rgb]{0.00,0.00,0.81}{#1}}

\newcommand{\ExtensionTok}[1]{#1}

\newcommand{\FunctionTok}[1]{\textcolor[rgb]{0.13,0.29,0.53}{\textbf{#1}}}

\newcommand{\KeywordTok}[1]{\textcolor[rgb]{0.13,0.29,0.53}{\textbf{#1}}}
\newcommand{\NormalTok}[1]{#1}

\newcommand{\StringTok}[1]{\textcolor[rgb]{0.31,0.60,0.02}{#1}}
\newcommand{\VariableTok}[1]{\textcolor[rgb]{0.00,0.00,0.00}{#1}}

\usepackage{longtable,booktabs,array}
\usepackage{calc} 
\usepackage{etoolbox}
\makeatletter
\patchcmd\longtable{\par}{\if@noskipsec\mbox{}\fi\par}{}{}
\makeatother
\IfFileExists{footnotehyper.sty}{\usepackage{footnotehyper}}{\usepackage{footnote}}
\makesavenoteenv{longtable}
\usepackage{graphicx}
\makeatletter
\newsavebox\pandoc@box
\newcommand*\pandocbounded[1]{
  \sbox\pandoc@box{#1}%
  \Gscale@div\@tempa{\textheight}{\dimexpr\ht\pandoc@box+\dp\pandoc@box\relax}%
  \Gscale@div\@tempb{\linewidth}{\wd\pandoc@box}%
  \ifdim\@tempb\p@<\@tempa\p@\let\@tempa\@tempb\fi
  \ifdim\@tempa\p@<\p@\scalebox{\@tempa}{\usebox\pandoc@box}%
  \else\usebox{\pandoc@box}%
  \fi%
}
\def\fps@figure{htbp}
\makeatother
\providecommand{\tightlist}{%
  \setlength{\itemsep}{0pt}\setlength{\parskip}{0pt}}
\usepackage[]{natbib}
\usepackage{amsmath}
\usepackage{amsfonts}
\usepackage{graphicx}
\usepackage{booktabs}
\usepackage{tabularx}
\usepackage{subcaption}
\usepackage{bookmark}
\IfFileExists{xurl.sty}{\usepackage{xurl}}{} 
\hypersetup{
  pdftitle={abx\_amr\_simulator: A simulation environment for antibiotic prescribing policy optimization under antimicrobial resistance},
  pdfauthor={Joyce Lee; Seth Blumberg},
  pdfkeywords={healthcare, reinforcement learning, antibiotic resistance, antimicrobial resistance, simulation, optimization},
  hidelinks,
  pdfcreator={LaTeX via pandoc}}

\title{\texttt{abx\_amr\_simulator}: A simulation environment for antibiotic prescribing policy optimization under antimicrobial resistance}
\author{Joyce Lee \and Seth Blumberg}
\date{2026-03-11}

\begin{document}
\maketitle
\begin{abstract}
Antimicrobial resistance (AMR) poses a global health threat, reducing the effectiveness of antibiotics and complicating clinical decision-making. To address this challenge, we introduce \texttt{abx\_amr\_simulator}, a Python-based simulation package designed to model antibiotic prescribing and AMR dynamics within a controlled, reinforcement learning (RL)-compatible environment. The simulator allows users to specify patient populations, antibiotic-specific AMR response curves, and reward functions that balance immediate clinical benefit against long-term resistance management. Key features include a modular design for configuring patient attributes, antibiotic resistance dynamics modeled via a leaky-balloon abstraction, and tools to explore partial observability through noise, bias, and delay in observations. The package is compatible with the Gymnasium RL API, enabling users to train and test RL agents under diverse clinical scenarios. From an ML perspective, the package provides a configurable benchmark environment for sequential decision-making under uncertainty, including partial observability induced by noisy, biased, and delayed observations. By providing a customizable and extensible framework, \texttt{abx\_amr\_simulator} offers a valuable tool for studying AMR dynamics and optimizing antibiotic stewardship strategies under realistic uncertainty.
\end{abstract}

\section{Motivation and Significance}\label{motivation-and-significance}

Antimicrobial resistance (AMR) is widely recognized as a major threat to global public health, reducing the effectiveness of antibiotics and adversely affecting clinical outcomes worldwide. In 2019, an estimated 4.95 million deaths were associated with antibiotic-resistant bacterial infections, disproportionately affecting resource-limited settings \citep{who2015global}. In response, interventions such as antibiotic stewardship programs (ASPs) have been established to optimize antibiotic use and slow the spread of AMR; these programs consist of multidisciplinary efforts, i.e.~publishing antibiograms, clinician education, and prospective review of antibiotic regimens \citep{sutton2024antimicrobial}.

Despite their widespread adoption, quantitative evidence evaluating the population-level impact of ASPs remains limited. Existing studies are often constrained in scope or duration \citep{bertollo2018antimicrobial}, and key system components are incompletely observed. For example, antibiotic exposure is influenced by unmeasured factors such as agricultural use or environmental contamination, while there is no universally accepted way to measure community antimicrobial resistance levels. These challenges make it difficult to directly quantify long-term effects of prescribing interventions using observational studies alone.

To address this gap, simulation-based approaches offer a complementary framework for investigating AMR dynamics. In this study, we introduce \texttt{abx\_amr\_simulator}, a Python-based simulation environment for learning and evaluating prescribing policies under uncertainty \citep{lee2026abx_amr_simulator_package}. The simulator is structured as a configurable MDP/POMDP-style environment, with explicit control over observability (noise, bias, and delay), patient heterogeneity, and resistance dynamics. It is compatible with the Gymnasium reinforcement learning API \citep{brockman2016openai, towers2024gymnasium}, enabling standardized training and benchmarking of RL agents in settings with delayed feedback and long-horizon credit assignment.

Unlike existing simulation-based models, which primarily operate at the pathogen level --- including microbial evolution and treatment-cycling models \citep{weaver2024reinforcement}, multilevel resistance simulations \citep{campos2019simulating}, and individual-based bacterial population models \citep{park2018simulation} --- \texttt{abx\_amr\_simulator} is designed as a policy-learning environment at the patient-decision level. Individual prescribing decisions are the action unit, while community-level AMR prevalence is represented in the state. This abstraction complements pathogen-level simulators by directly supporting ML evaluation of decision policies, including RL methods for sequential control under partial observability.

\section{Software Description}\label{software-description}

\subsection{Software Architecture}\label{software-architecture}

\subsubsection{Simulating antibiotic prescribing/antimicrobial resistance dynamics as a Markov Decision Process}\label{sec:abx_amr_mdp}

The \texttt{abx\_amr\_simulator} models antibiotic prescribing and antimicrobial resistance (AMR) dynamics as a Markov Decision Process (MDP), a framework for optimizing sequential decision-making under uncertainty. In this model, the system evolves through a state space ( S ), an action space ( A ), and a stochastic transition function ( P(s'\textbar s,a) ), which determines how the environment changes based on prescribing decisions. At each time step, an agent observes the current state (or an observation vector in the case of Partially Observable MDPs) and selects an action according to a policy. The environment responds by transitioning to a new state and providing a scalar reward ( R ), reflecting the quality of the action's outcome.

The software architecture is organized around the \texttt{ABXAMREnv} class, which serves as the parent environment for simulation scenarios. Within this environment, three primary components capture the dynamics of antibiotic prescribing and AMR:

\begin{itemize}
\tightlist
\item
  \texttt{PatientGenerator}: This class models a synthetic patient population with baseline infection risks and attributes. It generates a fixed number of patients at each time step, providing the agent with cases requiring treatment decisions.
\item
  \texttt{RewardCalculator}: This class calculates the scalar reward ( R ) associated with prescribing decisions. Rewards are based on both clinical success (e.g., effective treatment of infections) and community-level health objectives, such as minimizing resistance levels.
\item
  \texttt{AMR\_LeakyBalloon}: An array of \texttt{AMR\_LeakyBalloon} instances capture the current resistance level for each antibiotic in the simulation. These models record how community-level resistance evolves as a function of prescribing frequency.
\end{itemize}

Together, these components define the simulation environment or scenario within the \texttt{ABXAMREnv} class. Once an agent is introduced --- whether it be a reinforcement learning (RL) agent, heuristic policy, or rules-based model --- the agent and environment participate in an interaction loop where the agent selects actions, and the environment transitions states and computes rewards in response.

The simulator conforms to the Gymnasium API, providing a standardized interface for evaluating decision-making strategies. Users can initialize a simulation by instantiating the \texttt{PatientGenerator} and \texttt{RewardCalculator} classes, which then get passed into the \texttt{PatientGenerator} initializer along with configuration parameters for the antibiotics to be included in the environment. This modular design allows researchers to easily customize populations, reward structures, and resistance dynamics.

\begin{figure}
\includegraphics[width=1\linewidth]{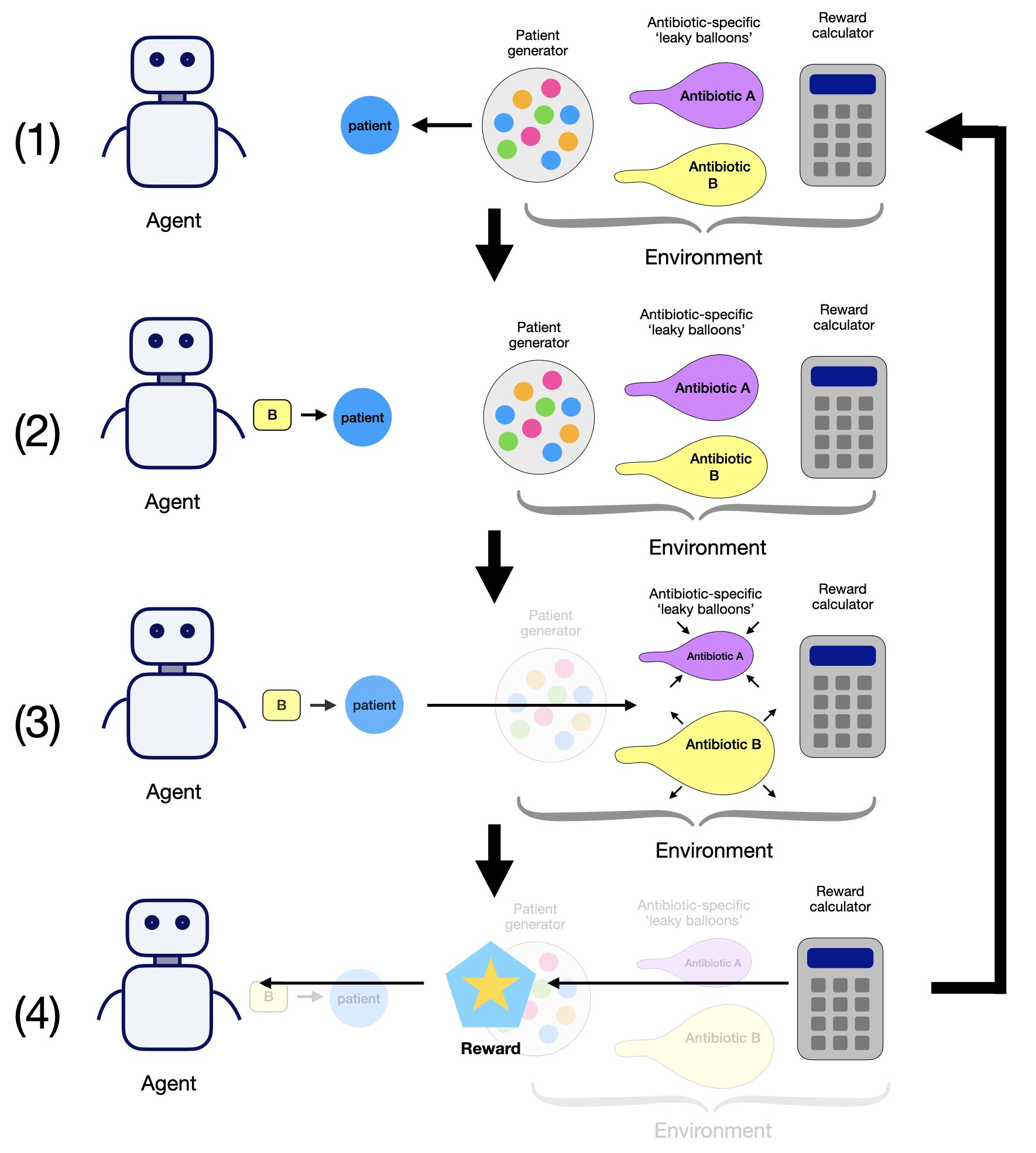} \caption{Schematic overview of the environment and its subcomponents (patient generator, antibiotic leaky balloons, and reward calculator), and the interaction loop between the agent and environment.}\label{fig:agent-env-interaction}
\end{figure}

Figure \ref{fig:agent-env-interaction} provides a schematic overview of the simulator's architecture and the agent-environment interaction loop.

\subsubsection{PatientGenerator: Synthetic Population Modeling}\label{patientgenerator-synthetic-population-modeling}

The \texttt{PatientGenerator} class creates a synthetic population of patients at each time step, with configurable attributes that define clinical profiles and expected treatment outcomes. Each patient is characterized by six attributes:

\begin{itemize}
\tightlist
\item
  Probability of infection (\(\pi_i\)): Likelihood of bacterial infection requiring treatment.
\item
  Benefit multipliers (\(\phi_{b,i}\), \(\omega_{b,i}\)): Scale the clinical benefit value and probability of successful treatment, respectively.
\item
  Failure multipliers (\(\phi_{f,i}\), \(\omega_{f,i}\)): Scale the penalty value and probability of treatment failure, respectively.
\item
  Recovery probability (\(\rho_i\)): Likelihood of spontaneous recovery without treatment.
\end{itemize}

These attributes can be configured for homogeneous or heterogeneous populations, with options to introduce noise or bias to observed values, enabling exploration of partial observability. At minimum, agents receive an estimated infection probability for each patient, while true infection status remains latent.

\subsubsection{AMR\_LeakyBalloon: Resistance Response Modeling}\label{amr_leakyballoon-resistance-response-modeling}

Antimicrobial resistance for each antibiotic is modeled using the \texttt{AMR\_LeakyBalloon} class, which implements a soft-bounded accumulator with decay dynamics. Resistance pressure increases with prescribing frequency and decays over time in the absence of selection pressure. Observable AMR levels (\(\sigma_a\)) are derived from latent resistance pressure via a sigmoid function, representing the probability that a new infection at the current time is resistant to a given antibiotic.

Each antibiotic is parameterized by:

\begin{itemize}
\tightlist
\item
  Flatness Parameter: Controls the steepness of resistance emergence.
\item
  Leak Parameter: Governs the rate of resistance decay.
\end{itemize}

The simulator also supports modeling cross-resistance, allowing the user to specify how prescribing one antibiotic influences resistance to others.

\subsubsection{RewardCalculator: Balancing Individual and Community Outcomes}\label{rewardcalculator-balancing-individual-and-community-outcomes}

The \texttt{RewardCalculator} class defines the scalar reward function for agent decisions, balancing immediate clinical success against long-term AMR outcomes. The overall reward function is split into two components:

\begin{itemize}
\tightlist
\item
  Individual Reward: Captures the net clinical benefit for each treated patient.
\item
  Community Reward: Encourages agents to minimize AMR levels across all antibiotics.
\end{itemize}

The reward function at time step \(t\) is defined as:

\begin{equation}
R_{\text{overall}}(t) = \left( 1 - \lambda \right) \cdot \langle \tilde{R}_{\text{individual}}(t) \rangle + \lambda \cdot \langle R_{\text{community}}(t) \rangle
\end{equation}

where \(\lambda \in [0, 1]\) allows users to adjust the weight between individual and community objectives. The reward function is modular, allowing users to adjust the weight of individual vs.~community objectives, and also to specifically tailor reward parameters for the individual objective as they wish.

For full derivations and parameter definitions, see tutorials and documentation available on the package's GitHub page \citep{lee2026abx_amr_simulator_package}.

\subsection{Software Functionalities}\label{software-functionalities}

\subsubsection{Flexible Simulation Environment for AMR Research}\label{flexible-simulation-environment-for-amr-research}

The \texttt{abx\_amr\_simulator} provides users with a modular Python framework for simulating antibiotic prescribing and antimicrobial resistance (AMR) dynamics. Its flexible design allows users to:

\begin{itemize}
\tightlist
\item
  Model synthetic patient populations with configurable attributes and partial observability (via the \texttt{PatientGenerator}).
\item
  Define antibiotic-specific resistance dynamics, including cross-resistance effects (via the \texttt{AMR\_LeakyBalloon}).
\item
  Balance short-term clinical success and long-term AMR management through customizable reward functions (via the \texttt{RewardCalculator}).
\item
  Explore complex prescribing scenarios by combining these components into a simulation environment (\texttt{ABXAMREnv}).
\end{itemize}

\subsubsection{Support for Reinforcement Learning Agents}\label{support-for-reinforcement-learning-agents}

While the simulator can accommodate various agent types, its primary functionality supports reinforcement learning (RL) agents for studying prescribing strategies. Users can:

\begin{itemize}
\tightlist
\item
  Test diverse RL algorithms, including advanced agents such as hierarchical RL agents.
\item
  Investigate how agents respond to varying levels of partial observability in patient attributes and resistance data.
\end{itemize}

\subsubsection{Automated Hyperparameter Tuning}\label{automated-hyperparameter-tuning}

The package integrates \texttt{Optuna} for automated hyperparameter tuning, allowing users to optimize agent and environment parameters \citep{akiba2019optuna}. Users can:

\begin{itemize}
\tightlist
\item
  Define tuning parameters via configuration files.
\item
  Conduct end-to-end pipelines where agents are tuned before full training episodes.
\end{itemize}

\subsubsection{Modular Configuration System}\label{modular-configuration-system}

The simulator uses YAML-based configuration files to allow users to easily modify and test different simulation setups without altering the codebase. Key features include:

\begin{itemize}
\tightlist
\item
  A hierarchical configuration system with subconfig files for each major class (PatientGenerator, AMR\_LeakyBalloon, RewardCalculator, etc.).
\item
  An umbrella configuration file that ties together subconfigs to define a complete experiment.
\end{itemize}

\subsubsection{Advanced Usage}\label{advanced-usage}

For users with more expertise, \texttt{abx\_amr\_simulator} supports creating custom subclasses for components such as \texttt{PatientGenerator}, \texttt{RewardCalculator}, or \texttt{AMR\_LeakyBalloon}. This allows researchers to adapt the simulator to novel scenarios or requirements. Tutorials on creating custom subclasses are available on the package's GitHub page \citep{lee2026abx_amr_simulator_package}.

\subsubsection{Graphical User Interface}\label{graphical-user-interface}

The \texttt{abx\_amr\_simulator} package includes a basic GUI built using the Streamlit library, providing an accessible interface for users to create and run experiments, and view diagnostic plots from completed experiments. The GUI enables users to:

\begin{itemize}
\tightlist
\item
  Configure simulation environments and prescribing scenarios without modifying YAML files directly.
\item
  Visualize key results from experiments, including the cumulative clinical outcomes of the treatment decisions made by the agent.
\end{itemize}

The GUI complements the command-line interface and YAML-based configuration system, offering a streamlined workflow for users who prefer graphical interaction.

\section{Illustrative Examples}\label{illustrative-examples}

\subsection{Setting Up and Running Experiments}\label{setting-up-and-running-experiments}

The \texttt{abx\_amr\_simulator} is a Python-based simulation framework designed for ease of use through YAML configuration files and CLI commands. Below, we describe the workflow for setting up a basic reinforcement learning (RL) experiment, highlighting key steps such as installation, configuration, and training.

\subsection{Installation and Setup}\label{installation-and-setup}

To begin, users must install \texttt{abx\_amr\_simulator} and its dependencies. The following commands create a Python environment, clone the repository, and install the package in editable mode:

\begin{Shaded}
\begin{Highlighting}[]
\ExtensionTok{conda}\NormalTok{ create }\AttributeTok{{-}{-}name}\NormalTok{ abx\_amr python=3.10}
\FunctionTok{git}\NormalTok{ clone https://github.com/jl56923/abx\_amr\_simulator.git}
\BuiltInTok{cd}\NormalTok{ abx\_amr\_simulator}
\ExtensionTok{pip}\NormalTok{ install }\AttributeTok{{-}e}\NormalTok{ .}
\end{Highlighting}
\end{Shaded}

\subsection{Configuring Experiments}\label{configuring-experiments}

abx\_amr\_simulator uses YAML configuration files to simplify customization and experimentation. Users can generate default configuration files by running the setup\_config\_folders\_with\_defaults utility:

\begin{Shaded}
\begin{Highlighting}[]
\FunctionTok{mkdir}\NormalTok{ abx\_amr\_project}
\BuiltInTok{cd}\NormalTok{ abx\_amr\_project}
\FunctionTok{mkdir}\NormalTok{ experiments}
\BuiltInTok{cd}\NormalTok{ experiments}

\ExtensionTok{python} \AttributeTok{{-}c} \StringTok{"from abx\_amr\_simulator.utils import setup\_config\_folders\_with\_defaults; \textbackslash{} }
\StringTok{  from pathlib import Path; setup\_config\_folders\_with\_defaults(Path(\textquotesingle{}.\textquotesingle{}))"}
\end{Highlighting}
\end{Shaded}

This command creates a configs/ directory containing modular configuration files for each component of the simulation:

\begin{itemize}
\tightlist
\item
  Environment Configuration (configs/environment/default.yaml): Specifies the patient population, infection dynamics, and simulation parameters.
\item
  Reward Function Configuration (configs/reward\_calculator/default.yaml): Defines how rewards are calculated based on antibiotic prescribing decisions.
\item
  Patient Generator Configuration (configs/patient\_generator/default.yaml): Controls the distribution of patient characteristics and infection risk.
\item
  Agent Algorithm Configuration (configs/agent\_algorithm/default.yaml): Includes hyperparameters for training RL agents, such as those used in PPO.
\end{itemize}

The modular structure allows users to adapt individual components without affecting the overall experiment. For example, the configs/umbrella\_configs/base\_experiment.yaml file acts as the central configuration, referencing subconfigs and defining training parameters:

\begin{Shaded}
\begin{Highlighting}[]
\CommentTok{\# Base experiment configuration}
\FunctionTok{config\_folder\_location}\KeywordTok{:}\AttributeTok{ ../}
\FunctionTok{options\_folder\_location}\KeywordTok{:}\AttributeTok{ ../../options}

\CommentTok{\# Component configurations}
\FunctionTok{environment}\KeywordTok{:}\AttributeTok{ environment/default.yaml}
\FunctionTok{reward\_calculator}\KeywordTok{:}\AttributeTok{ reward\_calculator/default.yaml}
\FunctionTok{patient\_generator}\KeywordTok{:}\AttributeTok{ patient\_generator/default.yaml}
\FunctionTok{agent\_algorithm}\KeywordTok{:}\AttributeTok{ agent\_algorithm/default.yaml}

\CommentTok{\# Training configuration}
\FunctionTok{training}\KeywordTok{:}
\AttributeTok{  }\FunctionTok{run\_name}\KeywordTok{:}\AttributeTok{ example\_run}
\AttributeTok{  }\FunctionTok{total\_num\_training\_episodes}\KeywordTok{:}\AttributeTok{ }\DecValTok{25}
\AttributeTok{  }\FunctionTok{save\_freq\_every\_n\_episodes}\KeywordTok{:}\AttributeTok{ }\DecValTok{5}
\AttributeTok{  }\FunctionTok{eval\_freq\_every\_n\_episodes}\KeywordTok{:}\AttributeTok{ }\DecValTok{5}
\AttributeTok{  }\FunctionTok{num\_eval\_episodes}\KeywordTok{:}\AttributeTok{ }\DecValTok{10}
\AttributeTok{  }\FunctionTok{seed}\KeywordTok{:}\AttributeTok{ }\DecValTok{42}
\AttributeTok{  }\FunctionTok{log\_patient\_trajectories}\KeywordTok{:}\AttributeTok{ }\CharTok{true}
\end{Highlighting}
\end{Shaded}

Training episodes are defined by the max\_time\_steps parameter in the environment subconfig, ensuring consistent episode length. This design reflects the simulator's simplified approach, where episodes terminate only after the maximum time steps are reached.

\section{Training an RL Agent}\label{training-an-rl-agent}

Once the configuration files are set up, users can train an RL agent using the following CLI command:

\begin{Shaded}
\begin{Highlighting}[]
\ExtensionTok{python} \AttributeTok{{-}m}\NormalTok{ abx\_amr\_simulator.training.train }\DataTypeTok{\textbackslash{}}
  \AttributeTok{{-}{-}config}\NormalTok{ configs/umbrella\_configs/base\_experiment.yaml}
\end{Highlighting}
\end{Shaded}

During training, results are saved to a timestamped folder within the results/ directory (results/\_). The CLI also supports overriding subconfigurations or individual parameters directly:

\begin{Shaded}
\begin{Highlighting}[]
\ExtensionTok{python} \AttributeTok{{-}m}\NormalTok{ abx\_amr\_simulator.training.train }\DataTypeTok{\textbackslash{}}
  \AttributeTok{{-}{-}config}\NormalTok{ configs/umbrella\_configs/base\_experiment.yaml }\DataTypeTok{\textbackslash{}}
  \AttributeTok{{-}{-}s} \StringTok{"environment{-}subconfig=}\VariableTok{$PROJECT\_FOLDER}\StringTok{/experiments/configs/}\DataTypeTok{\textbackslash{}}
\StringTok{    environment/three\_abx\_w\_crossresistance.yaml"} \DataTypeTok{\textbackslash{}}
  \AttributeTok{{-}{-}p} \StringTok{"environment.num\_patients\_per\_time\_step=5"} \DataTypeTok{\textbackslash{}}
  \AttributeTok{{-}{-}p} \StringTok{"training.run\_name=three\_abx\_w\_crossr\_exp"}
\end{Highlighting}
\end{Shaded}

This flexibility enables users to explore complex scenarios, such as cross-resistance or varying patient populations, by customizing parameters dynamically.

\subsection{Visualizing Results}\label{visualizing-results}

The \texttt{abx\_amr\_simulator} package provides two lightweight Streamlit GUIs intended for rapid, accessible workflows: an Experiment Runner for configuring and launching basic training runs, and an Experiment Viewer for browsing completed runs and inspecting saved configurations and diagnostic plots. These interfaces are designed to lower the barrier to entry for exploratory use and quick iteration, while preserving reproducibility through generated config files and standardized results folders.

\begin{figure}
\includegraphics[width=0.9\linewidth]{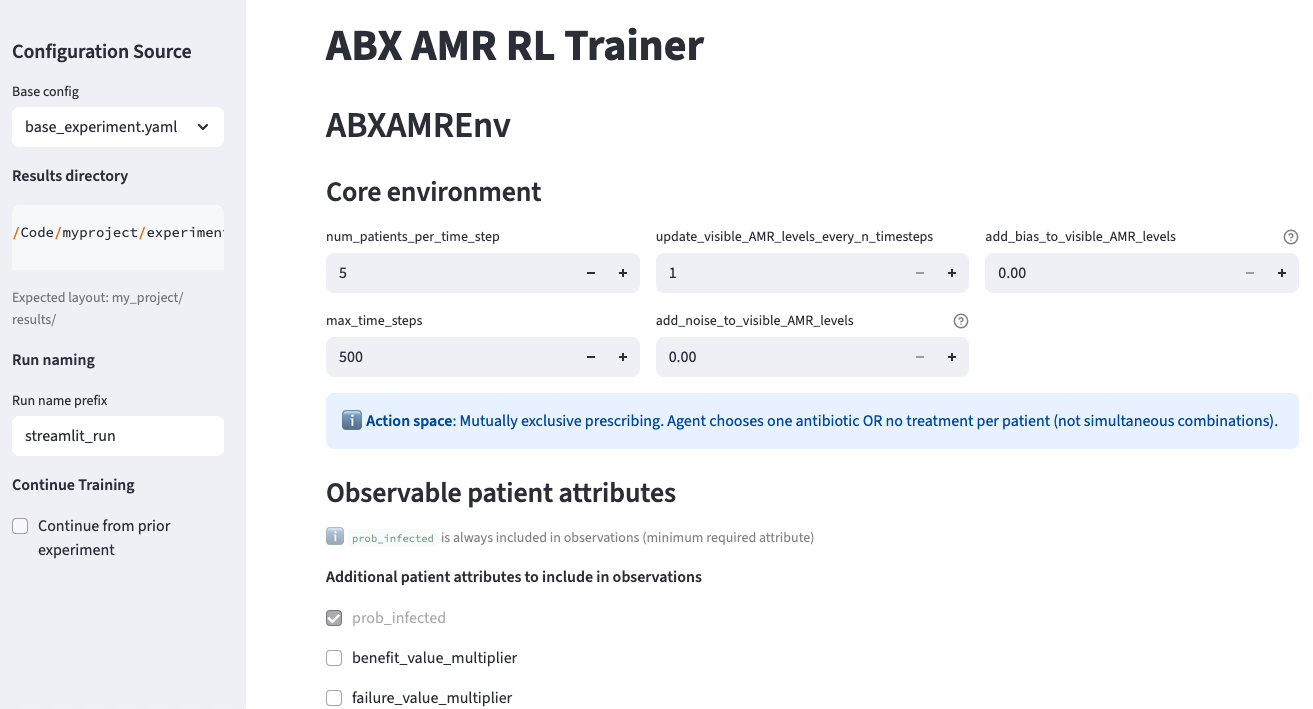} \caption{Experiment Runner GUI to run lightweight RL experiments}\label{fig:abx-amr-simulator-experiment-runner}
\end{figure}

\begin{figure}
\includegraphics[width=0.9\linewidth]{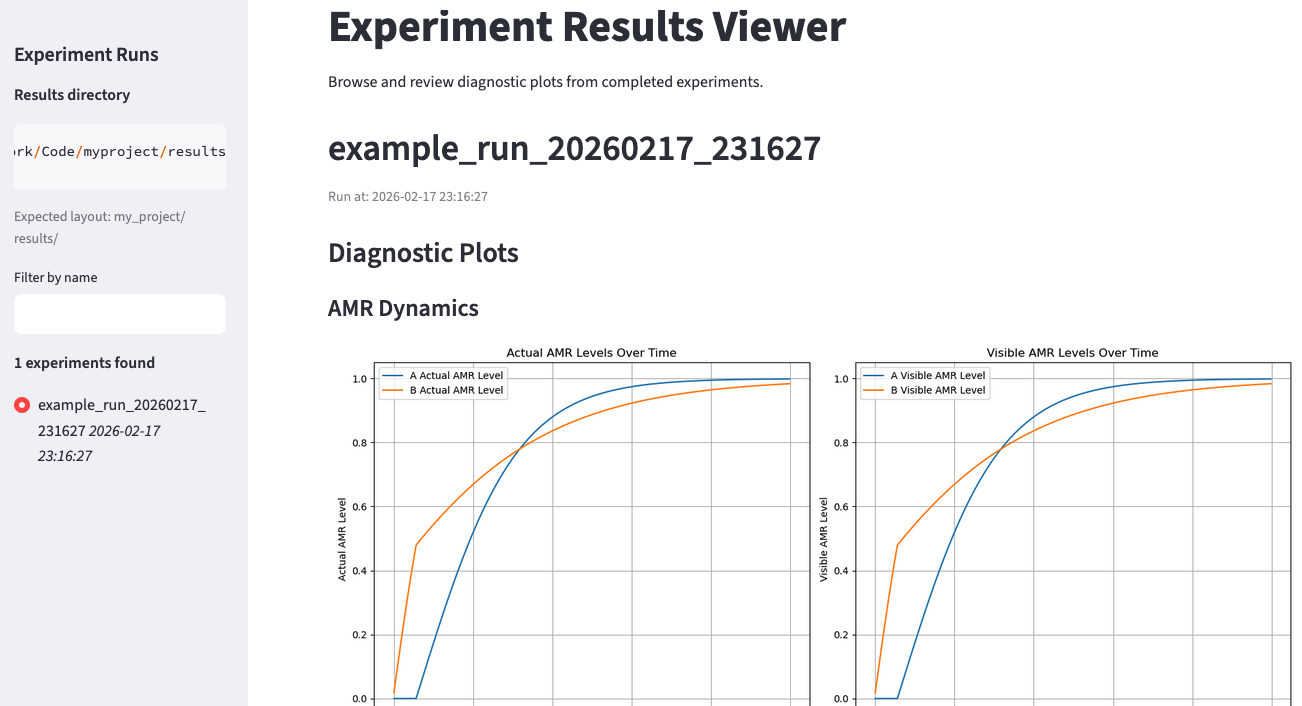} \caption{Experiment Viewer GUI to view results of RL experiments}\label{fig:abx-amr-simulator-experiment-viewer}
\end{figure}

\section{Impact}\label{impact}

The \texttt{abx\_amr\_simulator} package addresses a critical gap in antimicrobial resistance (AMR) research by providing a quantitative framework for evaluating antibiotic prescribing policies under varying levels of partial observability. By enabling controlled exploration of scenarios such as delayed access to resistance information (e.g., infrequently updated antibiograms) and heterogeneous patient populations subject to biased or noisy risk assessments, the simulator facilitates systematic analysis of how uncertainty and information degradation influence optimal prescribing behavior.

This modular simulation environment allows researchers to investigate key public health questions, such as:

\begin{itemize}
\tightlist
\item
  How do different clinical or informational constraints shape prescribing strategies?
\item
  What are the quantitative benefits of interventions like more frequent antibiogram updates or improved patient risk assessments?
\end{itemize}

The simulator serves as a ``quantitative laboratory'' for the public health and epidemiology communities. It enables researchers to disentangle how specific data deficiencies impede antibiotic stewardship efforts and to assess the potential gains achievable through targeted interventions.

Beyond its current functionality, \texttt{abx\_amr\_simulator} has significant potential for future expansion. Planned developments include support for nonstationary dynamics in both patient populations and resistance mechanisms. Introducing nonstationary behavior would allow users to model:

\begin{itemize}
\tightlist
\item
  Demographic shifts, seasonal infection trends, and high-risk subpopulations.
\item
  Irreversible changes in resistance mechanisms, capturing phenomena where prolonged antibiotic use creates lasting effects on community-level resistance.
\end{itemize}

These extensions would enable researchers to address clinically relevant questions, such as how frequently prescribing guidelines should be updated, what signals indicate outdated policies, and how delayed policy revisions affect overall clinical outcomes for the community.

Additionally, we also plan to extend the simulator to support multi-agent, multi-locale experiments, where multiple agents operate across geographically distinct communities with unique resistance dynamics. Patients can migrate between locales, spreading resistant infections, while agents (representing individual clinicians or healthcare facilities) make prescribing decisions based only on local data. This extension will enable exploration of emergent behaviors, including:

\begin{itemize}
\tightlist
\item
  How localized prescribing strategies influence community-wide resistance.
\item
  Whether coordination mechanisms between agents improve collective outcomes.
\item
  How informational asymmetries across locales affect optimal prescribing policies.
\end{itemize}

Ultimately, \texttt{abx\_amr\_simulator} provides a robust platform for answering pressing research questions in AMR, offering public health and epidemiology researchers a tool to systematically evaluate the trade-offs between clinical outcomes and long-term resistance mitigation.

\section{Conclusions}\label{conclusions}

In this work, we introduced \texttt{abx\_amr\_simulator}, a Python-based simulation framework designed for researchers in computational public health and epidemiology. The simulator combines modularity, extensibility, and ease of use, enabling users to run sophisticated experiments by configuring YAML files and using the command-line interface, while also supporting advanced customization through subclass definitions for users seeking greater control over patient populations, resistance dynamics, or reward structures.

The simulator's ability to model antibiotic prescribing policies under varying levels of partial observability provides a novel approach to understanding how data deficiencies and uncertainty influence optimal decision-making. By enabling users to construct diverse scenarios, it offers valuable insights into the trade-offs between immediate clinical success and long-term antimicrobial resistance management.

Looking ahead, we plan to enhance \texttt{abx\_amr\_simulator} by introducing support for nonstationary dynamics and multi-agent, multi-locale experiments. Nonstationary dynamics will allow researchers to model evolving patient populations, resistance mechanisms, and shifting antibiotic response curves over time. Multi-agent, multi-locale functionality will enable users to investigate decentralized prescribing systems, where local agents operate independently but are connected through shared patient migration and resistance dynamics. These features will further extend the simulator's ability to address real-world AMR research questions and inform evidence-based interventions.

\bibliography{abx-amr-simulator}

\end{document}